\def\year{2018}\relax
\documentclass[letterpaper]{article} 
\usepackage{aaai18}  
\usepackage{times}  
\usepackage{helvet}  
\usepackage{courier}  
\usepackage{url}  
\usepackage{graphicx}  

\usepackage[usenames,dvipsnames]{color}
\usepackage{colortbl}
\definecolor{mygray}{gray}{.9}
\definecolor{mypink}{rgb}{.99,.91,.95}
\definecolor{mycyan}{cmyk}{.3,0,0,0}

\usepackage[utf8]{inputenc} 
\usepackage[T1]{fontenc}    
\usepackage{hyperref}       
\usepackage{url}            
\usepackage{booktabs}       
\usepackage{amsfonts}       
\usepackage{nicefrac}       
\usepackage{microtype}      

\usepackage{times}
\usepackage{graphicx}
\usepackage{epsfig}
\usepackage{amsmath}
\usepackage{amssymb}
\usepackage{booktabs}
\usepackage{fmtcount}
\usepackage{stfloats}
\usepackage{ dsfont }

\DeclareMathOperator*{\argmin}{argmin}
\usepackage{algorithm}
\usepackage{algorithmic}
\usepackage{setspace}

\usepackage{epsfig}
\usepackage{array}
\usepackage{float}
\usepackage{caption}
\usepackage{enumitem}
\usepackage{bbding}
\usepackage{setspace}
\begin{document}
%
\title{Meta Networks for Neural Style Transfer}
\author{Falong Shen\thanks{This work was done when Falong Shen was an Intern at 360 AI Institute.}\\
Peking University\\
shenfalong@pku.edu.cn
\And
Shuicheng Yan\\
360 AI Institute \\
National University of Singapore\\
yanshuicheng@360.cn
\And
Gang Zeng\\
Peking University\\
zeng@pku.edu.cn}
\maketitle
\begin{abstract}
In this paper we propose a new method to get the specified network parameters through one time feed-forward propagation of the meta networks and explore the application to neural style transfer. Recent works on style transfer typically need to train image transformation networks for every new style, and the style is encoded in the network parameters by enormous iterations of stochastic gradient descent. To tackle these issues, we build a meta network which takes in the style image and produces a corresponding image transformations network directly. Compared with optimization-based methods for every style, our meta networks can handle an arbitrary new style within $19ms$ seconds on one modern GPU card. The fast image transformation network generated by our meta network is only 449KB, which is capable of real-time executing on a mobile device. We also investigate the manifold of the style transfer networks by operating the hidden features from meta networks. Experiments have well validated the effectiveness of our method. Code and trained models has been released~\url{https://github.com/FalongShen/styletransfer}.
\end{abstract}
\vspace{-5mm}
\section{Introduction}
Style transfer is a long-standing problem that seeks to migrate a style from a reference style image to another input picture~\cite{efros2001image,efros1999texture,elad2016style,frigo2016split,heeger1995pyramid,kyprianidis2013state}. This task consists of two steps: (i) texture extraction from the style image and (ii) rendering the content image with the texture.
For the first step, lots of methods have been proposed to represent the texture, most of which exploit the deep mid-layer features from the pre-trained convolutional neural network (CNN)~\cite{gatys2015neural,li2017demystifying}. Gatys et al. used the second-order statistics between feature activations across all channels in their pioneer work~\cite{gatys2015neural}, while Li et al.~\cite{li2017demystifying} found the channel-wise feature statistics (\emph{e.g.}, mean and variance) are enough to give similar performance on representing the texture.
For the second step, Gatys et al.~\cite{gatys2015neural} used a gradient-descent-based method to find an optimal image, which minimizes the distance to both the content image and the style image. Despite the surprising success, their method needs thousands of iterations of gradient descent through large networks for a new input image.
Recently, Johnson et al.~\cite{johnson2016perceptual} proposed an image transformation network, solving the gradient-descent-based optimization problem by one feed-forward propagation under the condition of a fixed style image. The effectiveness of this method indicates that the texture information of a style image can be encoded in one convolutional network.
Several other works on image transformation networks have been proposed ever since the neural art transfer has emerged, but these pre-trained image transformation networks are limited to single~\cite{ulyanov2016texture,li2016precomputed} or several styles~\cite{dumoulin2017learned,li2017diversified,zhang2017multi}.
Given a new style image, the image transformation network has to be re-trained end-to-end by an enormous number of iterations of stochastic gradient descent (SGD) to encode the new texture, which limits its scalability to large numbers of styles.

Let us re-think how the image transformation network replaces the gradient-descent-based optimization. It uses a CNN to learn a direct-mapping from the input image to the near-optimal solution image~\cite{johnson2016perceptual}. To obtain such an image transformation network, we need to minimize the empirical loss on the training content images for a fixed style image, which can be solved by SGD~\cite{johnson2016perceptual}.

Then we would ask a question, ``\emph{Is SGD the only method to get the solution network for a new style image? }"

The answer is \emph{No}. As our target is to get a near-optimal network and the input is a style image, it is natural to build a direct mapping between the two domains. Instead of SGD, we propose to build a meta network which takes in the style image and produces the corresponding image transformation network.

\begin{figure*}
  \centering
  \includegraphics[width=1.5\columnwidth]{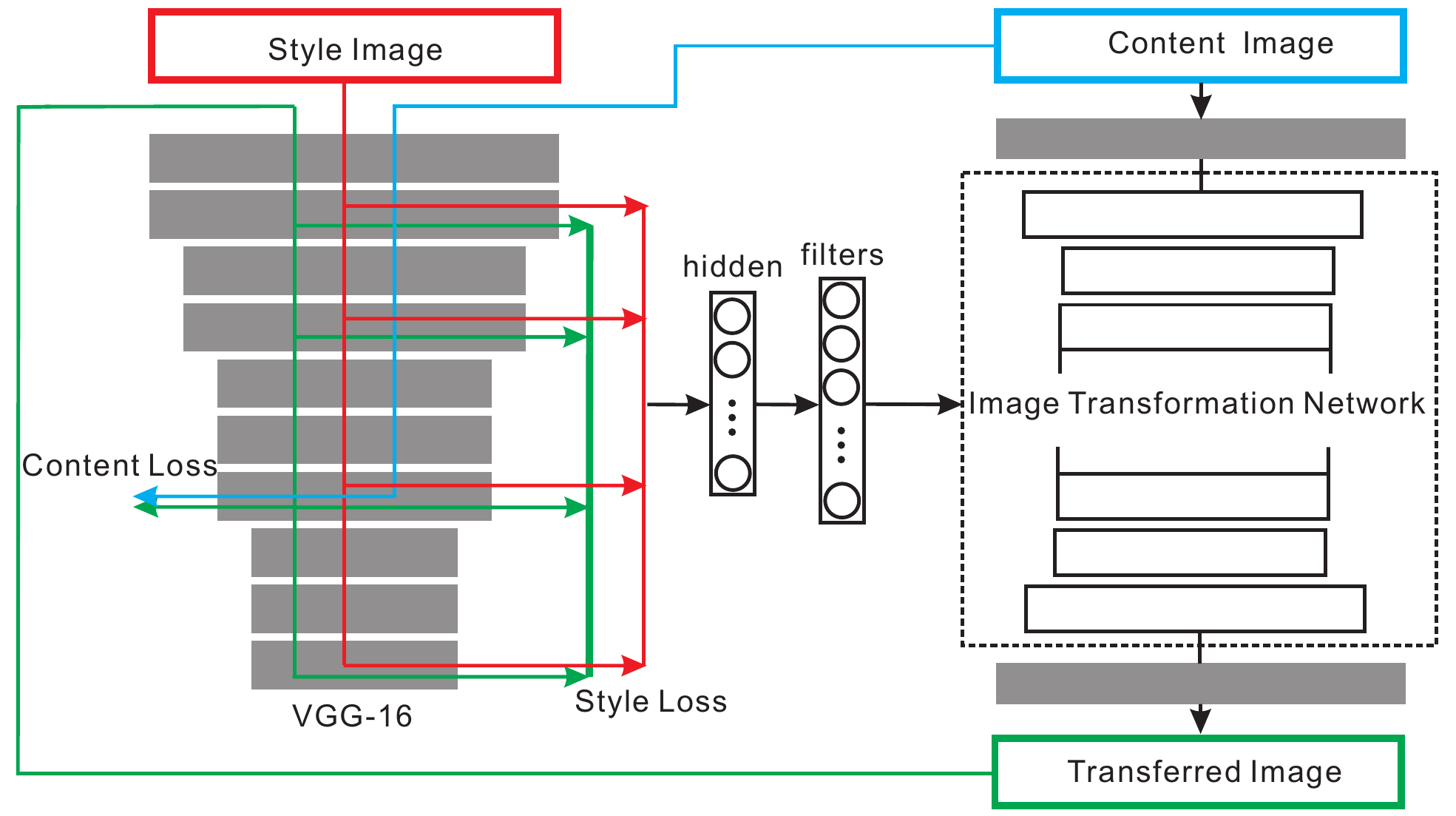}
  \small \caption{Model architecture. The style image is fed into the fixed VGG-16 to get the
  style feature, which goes through two fully connected layers to construct the filters for each \texttt{conv} layer in the corresponding image transformation network. We fixed the scale and bias in the instance \texttt{batchnorm} layer to $1$ and $0$. The dimension of hidden vector is $1792$ without specification. The hidden features are connected with the filters of each \texttt{conv} layer of the network in a group manner to decrease the parameter size, which means a 128 dimensional hidden vector for each \texttt{conv} layer.
  Then we compute the style loss and the content loss for the transferred image with the style image and the content image respectively through the fixed VGG-16.}
  \label{figure:metanetwork}
  \vspace{-5mm}
\end{figure*}

The meta network is composed of a frozen VGG-16 network~\cite{simonyan2014very} which extracts texture features from the given style image, and a series of fully connected layers to project the texture features to the transformation network space.
It is optimized by the empirical risk minimization across both the training content images and the style images. In this way, for the first time, we provide a new approach to generate an image transformation network for neural style transfer. We name it the ``meta network'' because it is able to produce different networks for different style images. The model architecture is depicted in Figure \ref{figure:metanetwork}.

The image transformation network is embedded as a hidden vector in the bottle-neck layer of the meta network. By interpolating the embedding hidden vectors of two networks induced from two real textures, we verify that the meta network generalizes the image textures rather than simply memorizing them.

The contributions of this paper are summarized as follows:
\begin{itemize}
  \item We address the network generation task and provide a meta network to generate networks for a specific domain. Specifically, the meta network takes in the new style image and produces a corresponding image transformation network in one feed-forward propagation.


  \item Our method provides an explicit representation of image transformation networks for neural style transfer, which enables texture synthesis and texture generation naturally.

  \item The generated networks from the meta network have the similar performance compared with SGD-based methods, but with orders of magnitude faster speed ($19ms$ v.s. $4h$) for a new style.

  \item We provide a new perspective on the algorithms for neural style transfer, which indicates that convolutional neural networks can be applied to optimization problems.
\end{itemize}

\vspace{-5mm}
\section{Related Work}
\paragraph{Hypernetworks and Meta Networks.}
A hypernetwork is a small network which is used to generate weights for a larger network. HyperNEAT~\cite{stanley2009hypercube} takes in a set of virtual coordinates to produce the weights.
Recently, Ha et al.~\cite{ha2016hypernetworks} proposed to use static hypernetworks to generate weights for a convolutional neural network and to use dynamic hypernetworks to generate weights for recurrent networks, where they took the hypernetwork as a relaxed form of weight sharing.

The works on meta networks adopt a two-level learning, a slow learning of a meta-level
model performing across tasks and a rapid learning of a base-level model acting within each task~\cite{mitchell1993explanation,vilalta2002perspective}.
Munkhdalai et al.~\cite{Munkhdalai2017metanetworks} proposed a kind of meta networks for one-shot classification via fast parameterization for the rapid generalization .

\paragraph{Style Transfer.}
Gatys et al.~\cite{gatys2015neural} for the first time proposed the combination of content loss and style loss based on the pre-trained neural networks on ImageNet~\cite{deng2009imagenet}. They approached the optimal solution image with hundreds of gradient descent iterations and produced high quality results. Then \cite{johnson2016perceptual} proposed to use image transformation networks to directly approach the near-optimal solution image instead of gradient descent. However, it needs to train an image transformation network for each new style, which is time consuming.

Recently, lots of improvements on~\cite{gatys2015neural} have been made. Instead of using the gram matrix of feature maps to represent the style, Li et al.~\cite{li2017demystifying} demonstrated that several other loss functions can also work, especially the mean-variance representation, which is much more compact than the gram matrix representation while giving similar performance. There are also other representations of the style, such as histogram loss~\cite{risser2017stable}, MRF loss~\cite{li2016combining} and CORAL loss~\cite{peng2017synthetic}.
Dumoulin et al.~\cite{dumoulin2017learned} proposed to use conditional instance normalization to accommodate each style. This method adjusts the weights of each channel of features and successfully represents many different styles. Unfortunately, it cannot be generalized to a new style image.
Huang et al.~\cite{huang2017arbitrary} found matching the mean-variance statistics of features from VGG-16 between the style image and the input image is enough to transfer the style. Although this method is able to process arbitrary new style, it heavily relies on a VGG-16 network to encode the image and also decode the feature by a corresponding  network, which makes it difficult to control the model size.
Chen et al.~\cite{chen2016fast} introduced a style swap layer to handle an arbitrary style transfer. Similar as the work~\cite{huang2017arbitrary}, they also proposed to adjust the feature in the content image towards the style image but in a patch-by-patch manner, which is much slower.

\vspace{-3mm}
\section{Proposed Method}
To find the optimal point of a function, gradient descent is typically adopted.
To find the optimal function for a specific task, the traditional method is to parameterize the function and optimize the loss function on the training data by SGD.
In this paper, we propose meta networks to find the near-optimal network directly, which will be detailed in this section. We also discuss its application to neural style transfer.
\subsection{Meta Networks}
\label{section:metanetworks}

Denote $f(x)$ and $h(x)$ as fixed differentiable functions and $||\cdot||$ is a norm. Let us consider an optimization problem of this kind:
\begin{equation}
    ||f(x)-f(a)|| + \lambda ||h(x)-h(b)||.
\end{equation}
There are three situations depending on whether $a$ and $b$ are fixed or variable.
\paragraph{Situation 1:} \emph{$a = a_0$ and $b = b_0$. Both are fixed.}

If $f(x)$ and $h(x)$ are convex functions, Eqn.~(1) is a typical convex optimization problem with regard to $x$ and we can apply gradient descent to obtain the optimal $x$ directly.
\begin{equation}
    \argmin\limits_x ||f(x)-f(a_0)|| + \lambda ||h(x)-h(b_0)||.
\end{equation}

\paragraph{Situation 2:} \emph{$b = b_0$. a is variable.}

For any given $a$, according to \textbf{Situation 1}, we have a corresponding $x$ which satisfies the function.
That is, there always exists a mapping function
\begin{equation}
    \mathcal{N}: a |\rightarrow x.
\end{equation}
We represent the mapping function as a deep neural network $\mathcal{N}(a;w)$ which is parameterized by $w$.
Now consider the \emph{empirical risk minimization} (ERM) problem to find the optimal mapping function:
\begin{equation}
    \argmin\limits_w \sum_a||f(x)-f(a)|| + \lambda ||h(x) - h(b_0)||
\end{equation}
where $x=\mathcal{N}(a;w)$.  However, it is not that trivial to find the optimal mapping function in this situation.
The main difficulty lies in optimization and the design of $\mathcal{N}(\cdot;w)$. The function $f(\cdot)$ is also important
for deciding whether a near-optimal mapping function can be approached by SGD.



\paragraph{Situation 3:} \emph{Both a and b are variable.}

For any given $b$, according to \textbf{Situation 2}, there exists a mapping function
which is able to find the near-optimal $x$ given $a$.
Suppose this optimal mapping function is parameterized by $w$ in $\mathcal{N}(\cdot;w)$
and can be approached by SGD.
Under this condition, there exists a mapping function:
\begin{equation}
    meta\mathcal{N}: b |\rightarrow \mathcal{N}(\cdot;w).
\end{equation}
Similar to \textbf{Situation 2}, we again represent this mapping function as a deep neural network $meta\mathcal{N}(b;\theta)$
which is parameterized by $\theta$. The meta network $meta\mathcal{N}(b;\theta)$ takes in $b$ as input
and produces a near-optimal network to Eqn. (2). Instead of iterations of SGD, the meta network needs only one
feed forward propagation to find the near-optimal network.

To optimize $\theta$ in the meta network, we consider the following ERM problem:
\begin{equation}
    \argmin\limits_{\theta} \sum_b\sum_a||f(x)-f(a)|| + \lambda ||h(x) - h(b)||,
\end{equation}
where $x=\mathcal{N}(a;w)$, and $w = meta\mathcal{N}(b;\theta)$. For every given $b$, there exists an optimal $w$, thus in the training stage it needs iterations of SGD to update the meta network parameter $\theta$
in order to produce an appropriate $w$ for each given $b$.

\begin{algorithm*}[!b]

\caption{Minibatch stochastic gradient descent training of meta networks for neural style transfer. We use $k=20$ and $m=8$ in our experiments. }

\quad \textbf{for} number of training iterations \textbf{do}

  \quad \quad $\vcenter{\hbox{\tiny$\bullet$}}$ Sample a style image $I_s$.

  \quad \quad \textbf{for} $k$ steps \textbf{do}

\begin{spacing}{1.0}
  \quad \quad \quad $\vcenter{\hbox{\tiny$\bullet$}}$   Feed-forward propagation of the meta network to get the transformation network
  \begin{center}
  $w \leftarrow meta\mathcal{N}(I_s;\theta)$.
  \end{center}
  \quad \quad \quad $\vcenter{\hbox{\tiny$\bullet$}}$  Sample minibatch of $m$ input images $\{I^{(1)}_c,...,I^{(m)}_c\}$.

  \quad \quad \quad $\vcenter{\hbox{\tiny$\bullet$}}$  Feed-forward propagation of the transformation network to get transferred images.

  \quad \quad \quad $\vcenter{\hbox{\tiny$\bullet$}}$ Computing the content loss and style loss and update $\theta$
  \begin{center}
   $\nabla_{\theta}\sum_{I_c}\bigg (\lambda_c||(\textbf{CP}(I)-\textbf{CP}(I_c))||_2^2 + \lambda_s||(\textbf{SP}(I)-\textbf{SP}(I_s))||_2^2\bigg )$
  \end{center}
  \quad \quad \textbf{end for}

\end{spacing}
\quad \textbf{end for}
\label{alogrithm:training}
\end{algorithm*}

\subsection{Neural Style Transfer}
Neural style transfer was first proposed by Gatys et al.~\cite{gatys2015neural} to render a content image in the style of another image
based on features extracted from a pre-trained deep neural network like VGG-16~\cite{simonyan2014very}. There are four situations depending on whether
the content image and the style image are fixed or variable.

\subsubsection{Fixed Content Image and Fixed Style Image}
For a pair of given images $(I_s,I_c)$, the target is to find an optimal image $I$ which minimizes the perceptual loss function to combine the style of $I_s$ and the content of $I_c$:
\begin{equation}\label{equation:Gatys}
\setlength{\jot}{-5pt}
\begin{split}
  \min\limits_{I}\bigg (\lambda_c||\textbf{CP}&(I;w_f)-\textbf{CP}(I_c;w_f)||_2^2 + \\
   &\lambda_s||\textbf{SP}(I;w_f)-\textbf{SP}(I_s;w_f)||_2^2 \bigg )
\end{split}
\end{equation}
where $\textbf{SP}(\cdot;w_f)$ and $\textbf{CP}(\cdot;w_f)$ are perceptual functions based on pre-trained deep neural networks parameterized by the fixed weights $w_f$,
which represent the style perceptron and the content perceptron individually.
$\lambda_s$ and $\lambda_c$ are scalars.
According to \textbf{Situation 1}, it is straightforward to apply gradient descent to the whole networks and use the gradient information from back-propagation to synthesize an image
to minimize the loss function.
This method produces high quality results for any given style image,
but needs hundreds of optimization iterations to get a converged result for each sample, which brings a large computation burden.
\subsubsection{Variable Content Image and Fixed Style Image}
\label{section:style2}
Instead of directly using the gradient information with regard to the input image to synthesize a new image, Johnson et al.~\cite{johnson2016perceptual} applied feed-forward image transformation to generate the target image. The image transformation networks are optimized across a large natural image dataset using the gradient of the parameters by back propagation:
\begin{equation}\label{equation:Gatys3}
\setlength{\jot}{-5pt}
\begin{split}
  \min\limits_{w}\sum_{I_c}\bigg (\lambda_c||\textbf{CP}&(I_w;w_f)-\textbf{CP}(I_c;w_f)||_2^2 + \\ &\lambda_s||\textbf{SP}(I_w;w_f)-\textbf{SP}(I_s;w_f)||_2^2 \bigg ),
\end{split}
\end{equation}
where $I_w = \mathcal{N}(I_c;w)$ and  $\mathcal{N}$ is the image transformation network which is parameterized by $w$. The style of $I_s$ is encoded in $w$. For a new content image, it only needs a forward propagation through the transformation network to generate the transferred image.

\subsubsection{Fixed Content Image and Variable Style Image}
As aforementioned in \textbf{Situation 2}, for some specified condition, there is a direct-mapping parameterized by CNN to approach the near-optimal solution for the loss. We consider the symmetry question. For a fixed content image, we try to find the transferred image for every style image and get the direct-mapping by
\begin{equation}\label{equation:Gatys2}
\setlength{\jot}{-5pt}
\begin{split}
  \min\limits_{w}\sum_{I_s}\bigg (\lambda_c||\textbf{CP}&(I_w;w_f)-\textbf{CP}(I_c;w_f)||_2^2 +\\ &\lambda_s||\textbf{SP}(I_w;w_f)-\textbf{SP}(I_s;w_f)||_2^2 \bigg ),
\end{split}
\end{equation}
where $I_w = \mathcal{N}(I_c;w)$. However, it has been found that it is not able to find the right mappings. The image transformation network only gives the style image as the transferred image, indicating that it is not that trivial to get a direct mapping to approach the gradient-descent solution.
\subsubsection{Variable Content Image and Variable Style Image}
To find a transformation network $\mathcal{N}(\cdot;w)$ which is parameterized by $w$ for a given style image $I_s$, it needs tens of thousands of iterations of SGD to get a satisfied network. As shown in \textbf{Situation 3}, we propose a natural way to get the transformation network $\mathcal{N}(\cdot;w)$ directly by a meta network:
\begin{equation}\label{equation:metanetworks}
  Meta\mathcal{N}: I_s  \rightarrow \mathcal{N}(\cdot;w).
\end{equation}
The meta network is parameterized by $\theta$ and is optimized across a dataset of content images and a dataset of style images  by
\begin{equation}\label{equation:Gatys4}
\setlength{\jot}{-5pt}
\begin{split}
  \min\limits_{\theta}\sum_{I_c,I_s}\bigg (\lambda_c||&(\textbf{CP}(I_{w_\theta};w_f)-\textbf{CP}(I_c;w_f))||_2^2 +\\ &\lambda_s||(\textbf{SP}(I_{w_\theta};w_f)-\textbf{SP}(I_s;w_f))||_2^2\bigg ),
\end{split}
\end{equation}
where $I_{w_\theta} = \mathcal{N}(I_c;w_\theta)$ and $w_\theta=Meta\mathcal{N}(I_s;\theta)$. The style image $I_s$ is used as the supervised target in the loss fucntion as well as input features to the meta network, which means the meta network takes a style image as input and generates a network which is able to transfer the content image towards the style image. Algorithm \ref{alogrithm:training} details the training strategy.

\vspace{-3mm}
\section{Experiment}

\begin{figure*}
\centering
    \begin{tabular}{c@{~}c@{~}c@{~}c@{~}c@{~}c@{~}}
\includegraphics[width=0.34\columnwidth,height=0.3\columnwidth]{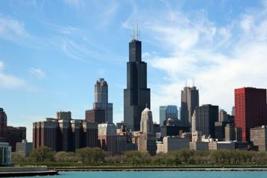}&
\includegraphics[width=0.34\columnwidth,height=0.3\columnwidth]{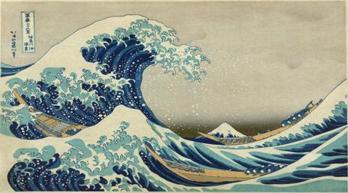}&
\includegraphics[width=0.34\columnwidth,height=0.3\columnwidth]{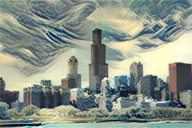}&
\includegraphics[width=0.34\columnwidth,height=0.3\columnwidth]{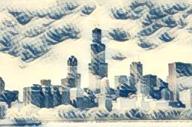}&
\includegraphics[width=0.34\columnwidth,height=0.3\columnwidth]{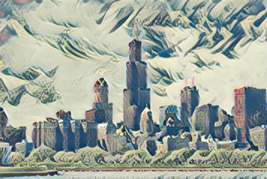}&
\includegraphics[width=0.34\columnwidth,height=0.3\columnwidth]{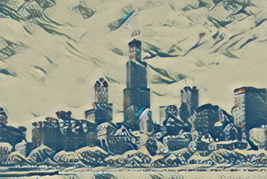}\\

\includegraphics[width=0.34\columnwidth,height=0.3\columnwidth]{figures/comparsions/stylecontent/chicago.jpg}&
\includegraphics[width=0.34\columnwidth,height=0.3\columnwidth]{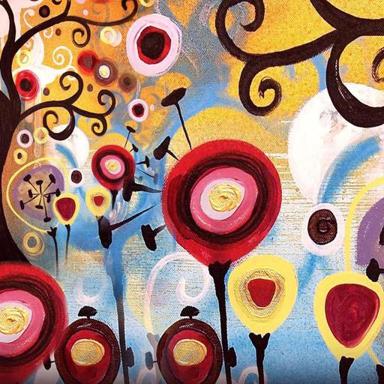}&
\includegraphics[width=0.34\columnwidth,height=0.3\columnwidth]{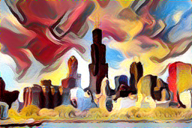}&
\includegraphics[width=0.34\columnwidth,height=0.3\columnwidth]{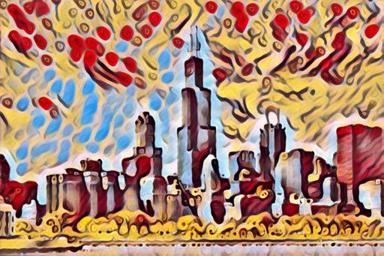}&
\includegraphics[width=0.34\columnwidth,height=0.3\columnwidth]{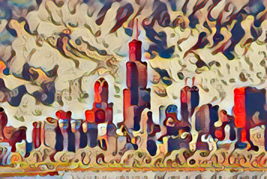}&
\includegraphics[width=0.34\columnwidth,height=0.3\columnwidth]{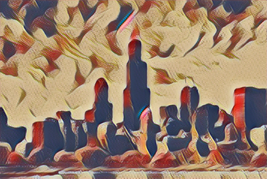}\\

\includegraphics[width=0.34\columnwidth,height=0.3\columnwidth]{figures/comparsions/stylecontent/chicago.jpg}&
\includegraphics[width=0.34\columnwidth,height=0.3\columnwidth]{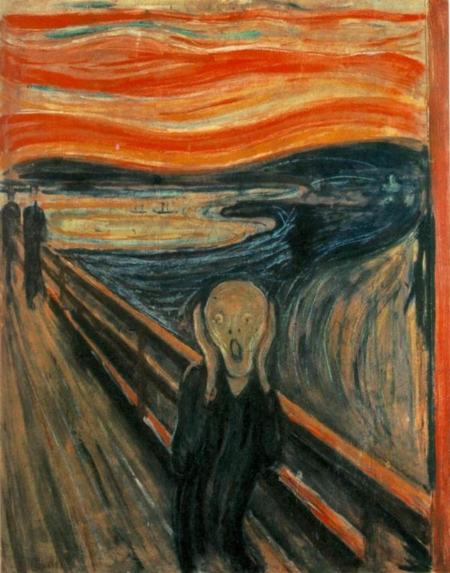}&
\includegraphics[width=0.34\columnwidth,height=0.3\columnwidth]{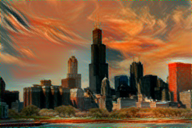}&
\includegraphics[width=0.34\columnwidth,height=0.3\columnwidth]{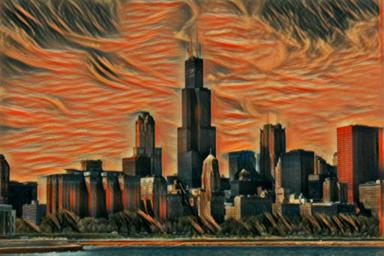}&
\includegraphics[width=0.34\columnwidth,height=0.3\columnwidth]{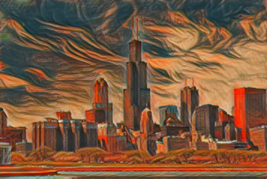}&
\includegraphics[width=0.34\columnwidth,height=0.3\columnwidth]{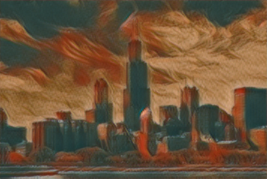}\\
(a) Input & (b) Style  & (c) Gatys \emph{et al.} & (d) John \emph{et al.}  & (e) Ours & (e) Ours-Fast
    \end{tabular}
    \small \caption{ Qualitative Comparisons of different methods. Both the style images and the content images are unseen in the \emph{train} set for \emph{our} model. }
    \label{figure:qulativecomparisions}
    \vspace{-7mm}
\end{figure*}

\begin{table*}[!b]
\vspace{-0mm}
\center
\small \caption{Comparison of different models. Our transfer network shares a similar structure with  but is much faster to encode the new style. The model in  can also be adapted to the new style efficiently at the price of relying on a VGG-16 network in the transferring stage. Both the style image and content image are resized to $256\times256$. Time is measured on a Pascal Titan X GPU.}
\begin{tabular}{p{5.0cm}<{\centering}|p{2cm}<{\centering}p{2cm}<{\centering}p{2cm}<{\centering}p{2cm}<{\centering}}
\toprule[2pt]
 Method              & \emph{Time(encode)}  &  \emph{Time(transfer)} & \emph{Size(transfer)} &  \emph{$\#$Style}\\
\midrule
\small{\cite{gatys2015neural}}                & N/A         & $9.52s$ & N/A & $\infty$ \\
\cite{johnson2016perceptual}              & $4h$          & $15ms$ & 7MB & 1\\
\cite{chen2016fast}                       &  $407ms$  & $171ms$ & 10MB& $\infty$\\
\cite{huang2017arbitrary}                       &  $27ms$  & $18ms$ & 25MB& $\infty$\\
\midrule
\textbf{Ours}                               & $19ms$     & $15ms$ &  7MB& $\infty$\\
\textbf{Ours-Fast}                               & $11ms$     & $8ms$ &  449KB& $\infty$\\
\bottomrule[2pt]
\end{tabular}
\label{table:comparisions}
\end{table*}

\subsection{Implementation Details}
 The meta networks for neural style transfer are trained on the content images from MS-COCO~\cite{lin2014microsoft} \emph{trainval} set and the style images  from the \emph{test} set of the WikiArt dataset~\cite{nicol2016wikiart}. There are about $120k$ images in MS-COCO \emph{trainval} set and about $80k$ images in the \emph{test} set of WikiArt. During training, each content image or style image is resized to keep the smallest dimension in the range $[256, 480]$, and randomly cropped regions of size $256\times256$. We use Adam \cite{kingma2014adam} with fixed learning rate 0.001 for $600k$ iterations without weight decay. The batch size of content images is 8 and the meta network is trained for $20$ iterations before changing the style image.
The transferred images are regularized with total variations loss with a strength of $10$. Our image transformation network shares the same structure with~\cite{johnson2016perceptual}, except that we remove the \texttt{Tanh} layer at last and the instance \texttt{BN} layer after the first \texttt{Conv} layer.

We compute the content loss at the $\tt{relu3\_3}$ layer and the style loss at layers $\tt{relu1\_2}$, $\tt{relu2\_2}$,$\tt{relu3\_3}$ and $\tt{relu4\_3}$ of VGG-16. The weight of content loss is $1$ while the weight of style loss is $250$. The content loss is the Euclidean distance between two feature maps of the content image and the transferred image. We compute the mean and stand deviations of two feature maps of the style image and the transferred image as style features.

\vspace{-3mm}
\subsection{Comparison with Other Methods}
We compare our method with other style transfer methods to evaluate its effectiveness and efficiency. Gatys et al.~\cite{gatys2015neural} proposed to find the optimal image by gradient descent and Johnson et al.~\cite{johnson2016perceptual} proposed to find the near-ptimal image transformation network by SGD. Compared with these two previous works, our meta network produces the image transformation network for each style by one forward-propagation. As the gradient-descent-based optimization and direct-mapping-based optimization share the same loss function, we make the comparison in terms of the converged loss, transferred image quality and running speed.

We show example style transfer results of different methods in Figure \ref{figure:qulativecomparisions}. Visually, there is no significantly difference between the three methods. Note that both the style image and the content image are unseen during the training stage of our meta networks while the model in~\cite{johnson2016perceptual} needs to be specifically trained end-to-end for every style. The image transformation network of our model and the model in~\cite{johnson2016perceptual} share the same architecture, our model can be easily generalized to any new style by only one forward propagation of the meta-network.
According to our experiments, it costs about $19ms$ to produce an image transformation network which is used in~\cite{johnson2016perceptual} for a single Titan X GPU card.

\begin{figure}
  \centering
  \includegraphics[width=0.8\columnwidth]{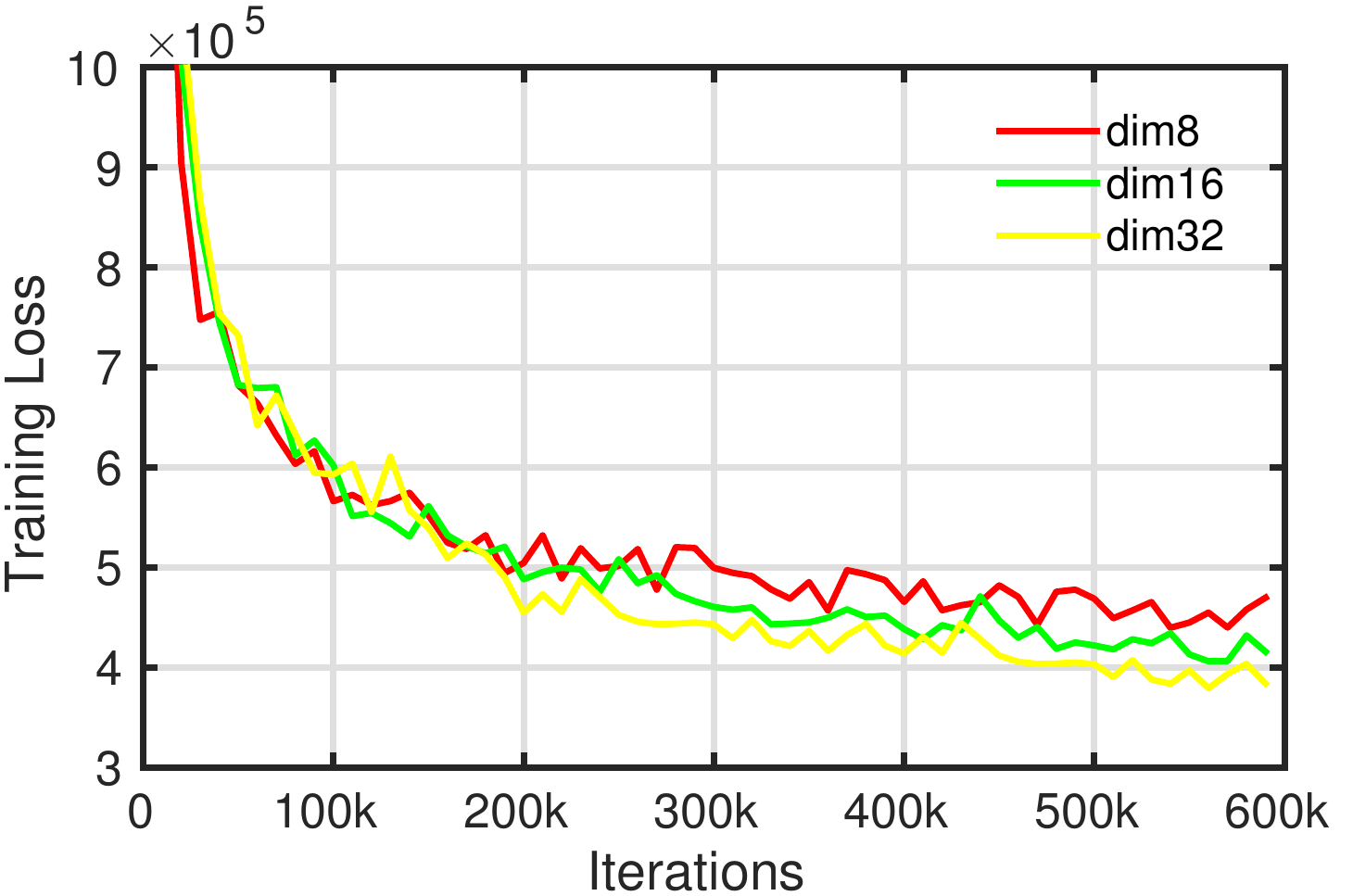}
  \small \caption{Training curves for image transformation networks of different model size. \texttt{dim32} denotes the filter numbers is 32 in the first \texttt{conv} layer.}
  \label{figure:traincurve}
\vspace{-7mm}
\end{figure}

Table \ref{table:comparisions} lists the advantages and defects of these methods. The gradient descent method of optimization in~\cite{gatys2015neural} is the most flexible but cannot process images in real time. The method of~\cite{huang2017arbitrary} is restricted by the VGG-16 network. Our method enjoys the flexibility while can process images in real time for both encoding the style and transferring images.
Figure \ref{figure:traincurve} plots the training curve for image transformation networks with different \texttt{conv} filter numbers. In our experiments, the converged losses on the training styles and testing styles are both around $4.0\times10^5$ for $dim32$, which equal approximately 200 steps of gradient descent in~\cite{gatys2015neural} (our own implementation).
\subsubsection{Difference with SGD solution.}
\begin{figure}

\centering
    \begin{tabular}{c@{~}|c@{~}c@{~}c@{~}c@{~}c@{~}}
\includegraphics[width=0.14\columnwidth,height=0.13\columnwidth]{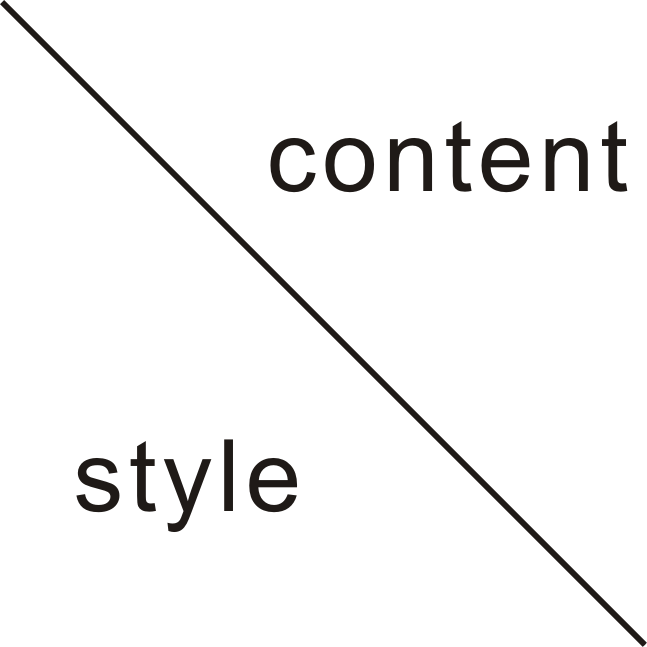}&
\includegraphics[width=0.14\columnwidth,height=0.13\columnwidth]{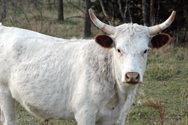}&
\includegraphics[width=0.14\columnwidth,height=0.13\columnwidth]{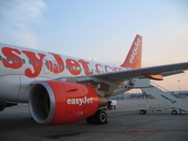}&
\includegraphics[width=0.14\columnwidth,height=0.13\columnwidth]{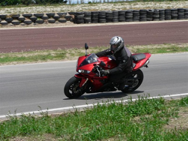}&
\includegraphics[width=0.14\columnwidth,height=0.13\columnwidth]{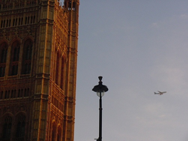}&
\includegraphics[width=0.14\columnwidth,height=0.13\columnwidth]{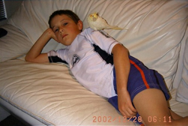}\\

\midrule

\includegraphics[width=0.14\columnwidth,height=0.13\columnwidth]{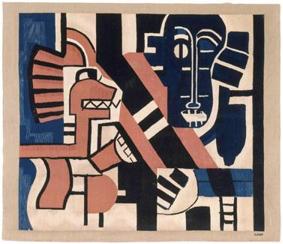}&
\includegraphics[width=0.14\columnwidth,height=0.13\columnwidth]{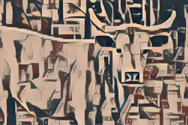}&
\includegraphics[width=0.14\columnwidth,height=0.13\columnwidth]{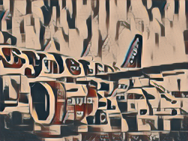}&
\includegraphics[width=0.14\columnwidth,height=0.13\columnwidth]{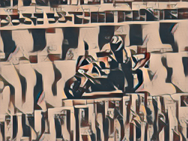}&
\includegraphics[width=0.14\columnwidth,height=0.13\columnwidth]{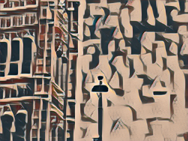}&
\includegraphics[width=0.14\columnwidth,height=0.13\columnwidth]{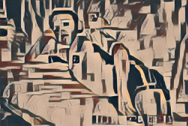}\\

\includegraphics[width=0.14\columnwidth,height=0.13\columnwidth]{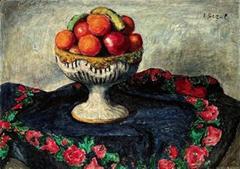}&
\includegraphics[width=0.14\columnwidth,height=0.13\columnwidth]{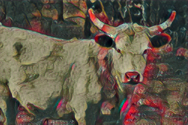}&
\includegraphics[width=0.14\columnwidth,height=0.13\columnwidth]{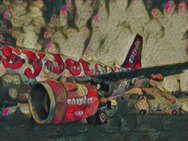}&
\includegraphics[width=0.14\columnwidth,height=0.13\columnwidth]{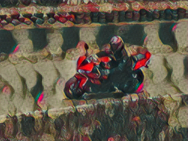}&
\includegraphics[width=0.14\columnwidth,height=0.13\columnwidth]{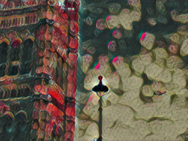}&
\includegraphics[width=0.14\columnwidth,height=0.13\columnwidth]{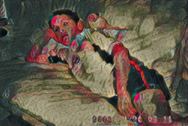}\\

\includegraphics[width=0.14\columnwidth,height=0.13\columnwidth]{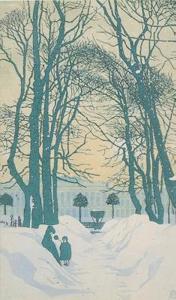}&
\includegraphics[width=0.14\columnwidth,height=0.13\columnwidth]{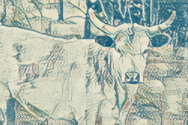}&
\includegraphics[width=0.14\columnwidth,height=0.13\columnwidth]{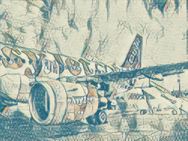}&
\includegraphics[width=0.14\columnwidth,height=0.13\columnwidth]{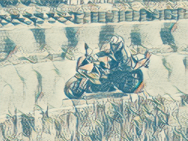}&
\includegraphics[width=0.14\columnwidth,height=0.13\columnwidth]{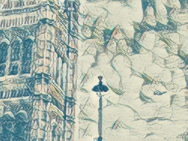}&
\includegraphics[width=0.14\columnwidth,height=0.13\columnwidth]{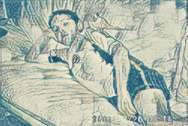}\\

\includegraphics[width=0.14\columnwidth,height=0.13\columnwidth]{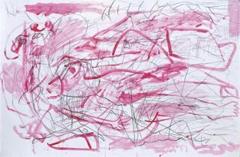}&
\includegraphics[width=0.14\columnwidth,height=0.13\columnwidth]{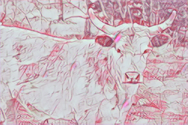}&
\includegraphics[width=0.14\columnwidth,height=0.13\columnwidth]{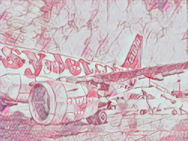}&
\includegraphics[width=0.14\columnwidth,height=0.13\columnwidth]{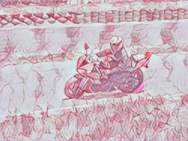}&
\includegraphics[width=0.14\columnwidth,height=0.13\columnwidth]{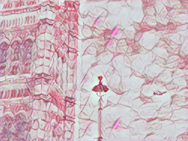}&
\includegraphics[width=0.14\columnwidth,height=0.13\columnwidth]{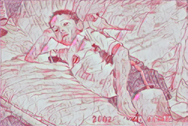}\\
    \end{tabular}
     \small \caption{ Examples of style transfer by the fast version of our meta networks. The size of image transformation network is 449KB, which is able to real-time execute on a mobile device. Best viewed in color.}
    \label{figure:goodfast}
    \vspace{-0mm}
\end{figure}

Comparing to \cite{johnson2016perceptual}, our meta network also takes in all the style features and generates almost the whole image transformation networks. Similar to SGD solution, our meta network has both the complete supervised information (all style features) and the total freedom (almost the whole image transformation networks). Therefore our image transformation networks can adopt any other useful structure, which is the same with SGD.
To show the effectiveness of the meta network, we have experimented with much smaller image transformation networks (449KB), the network architecture is characterized in Table \ref{table:fastnet}. This small and fast image transformation networks also give satisfying results as shown in Figure \ref{figure:goodfast} (best viewed in color).

\subsubsection{Difference with AdaBN solution.}
Recently, Huang et al.~\cite{huang2017arbitrary} also proposed an approach which can process an arbitrary style. They transfer images by adjusting the statistics (mean and std deviations) of the feature map in \texttt{AdaIN} layer and achieve surprisingly good results. However, their model needs to encode the image by fixed VGG-16 and decode the feature, which also requires a VGG-16-like architecture. This makes their image transformation network relatively large ($\sim$25MB).
The difference between this work and our method lies in the freedom of image transformation network for different style image. Huang et al.~\cite{huang2017arbitrary} aligns the mean and variance of the content feature maps to those of style feature maps, which means the style image only injects a 1024-dim vector to influence one layer in image transformation networks. The influence is quite limited because not all style features in supervised stage is used and only one layer is changed for different styles.

The key to success of the work in~\cite{huang2017arbitrary} is the high-level encoded feature from VGG-19. It should be noticed that they rely on VGG for both style image and image transformation networks. According to our experiments, their method fails when the pre-trained VGG-19 in image transformation networks is replaced. In the work by~\cite{huang2017arbitrary},  both the encoder (fixed to VGG) and decoder (trained) of image transformation networks is fixed except the \texttt{AdaIN} layer for any given style image. Although the image transformation networks is large ($\sim$25MB), it has little freedom for a new style image.


\begin{table}
\arrayrulecolor{white}
\centering
 \begin{tabular}{c@{~}|c@{~}}
   \hline\rowcolor{mycyan}layer & activation size \\
  \hline\rowcolor{mycyan}input & $3\times256\times256$ \\
  \hline\rowcolor{mycyan} Reflection Padding $(40\times40)$  &  $3\times336\times336$\\
  \hline\rowcolor{mygray}  $8\times9\times9$ conv, stride 1 & $8\times336\times336$\\
  \rowcolor{mypink}$16\times3\times3$ conv, stride 2 & $16\times168\times168$\\
  \rowcolor{mypink}$32\times3\times3$ conv, stride 2 & $32\times84\times84$\\
  \rowcolor{mypink}Residual block, 32 filters & $32\times80\times80$\\
  \rowcolor{mypink}Residual block, 32 filters & $32\times76\times76$\\
  \rowcolor{mypink}Residual block, 32 filters & $32\times72\times72$\\
  \rowcolor{mypink}Residual block, 32 filters & $32\times68\times68$\\
  \rowcolor{mypink}Residual block, 32 filters & $32\times64\times64$\\
  \rowcolor{mypink}$16\times3\times3$ deconv, stride 2 &  $16\times128\times128$\\
  \rowcolor{mypink}$8 \times3\times3$ deconv, stride 2 & $8\times256\times256$\\
   \hline\rowcolor{mygray}   $3\times9\times9$ conv, stride 1  & $3\times256\times256$\\
  \end{tabular}
  \small \caption{The fast version of image transformation networks. Every \texttt{conv} layer except for the first and the last is followed by a instance \texttt{batchnorm} layer and a \texttt{relu} layer sequentially, which are omitted in the table for clarity. All the filters of the \texttt{conv} layers in {\colorbox{mypink}{pink}}  region are generated by meta networks in the inference stage. The filters of the \texttt{conv} layers in {\colorbox{mygray}{gray}} region are jointly trained with meta networks and are fixed in the inference stage. There is no parameter in other layers. The model size is only 449KB, which is capable of executing on a mobile device.}
  \label{table:fastnet}
 \end{table}

\begin{figure}
\centering
    \begin{tabular}{c@{~}c@{~}c@{~}c@{~}c@{~}}
\includegraphics[width=0.18\columnwidth,height=0.15\columnwidth]{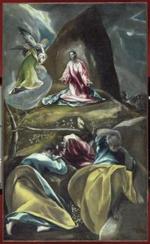}&
\includegraphics[width=0.18\columnwidth,height=0.15\columnwidth]{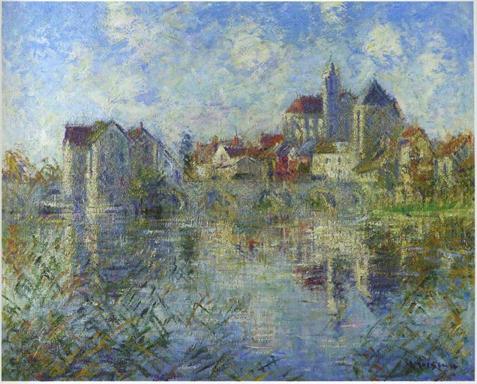}&
\includegraphics[width=0.18\columnwidth,height=0.15\columnwidth]{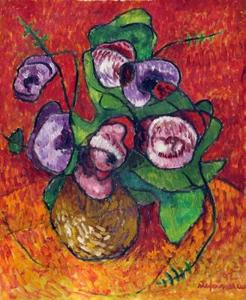}&
\includegraphics[width=0.18\columnwidth,height=0.15\columnwidth]{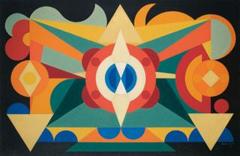}&
\includegraphics[width=0.18\columnwidth,height=0.15\columnwidth]{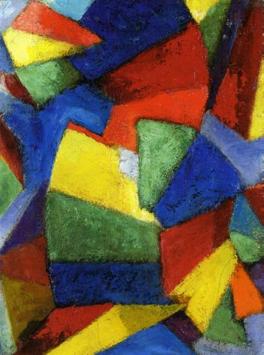}\\

\includegraphics[width=0.18\columnwidth,height=0.15\columnwidth]{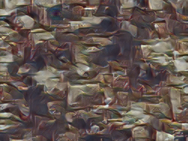}&
\includegraphics[width=0.18\columnwidth,height=0.15\columnwidth]{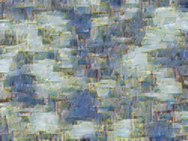}&
\includegraphics[width=0.18\columnwidth,height=0.15\columnwidth]{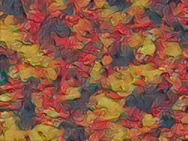}&
\includegraphics[width=0.18\columnwidth,height=0.15\columnwidth]{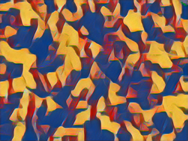}&
\includegraphics[width=0.18\columnwidth,height=0.15\columnwidth]{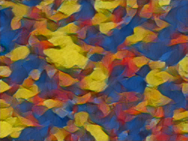}
    \end{tabular}
    \small \caption{ Texture Visualization. The first row displays style images and the second row displays the corresponding textures encoded in the image transformation networks.}
    \label{figure:texture}
\vspace{-5mm}
\end{figure}

\begin{figure*}[t]
\centering
    \begin{tabular}{c@{~}c@{~}c@{~}c@{~}c@{~}c@{~}}
\includegraphics[width=0.28\columnwidth,height=0.21\columnwidth]{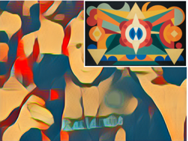}&
\includegraphics[width=0.28\columnwidth,height=0.21\columnwidth]{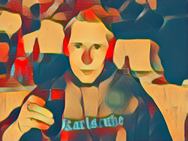}&
\includegraphics[width=0.28\columnwidth,height=0.21\columnwidth]{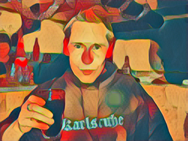}&
\includegraphics[width=0.28\columnwidth,height=0.21\columnwidth]{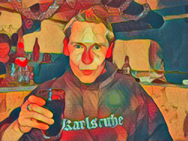}&
\includegraphics[width=0.28\columnwidth,height=0.21\columnwidth]{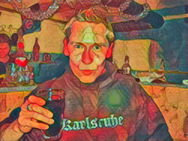}&
\includegraphics[width=0.28\columnwidth,height=0.21\columnwidth]{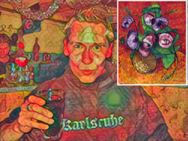}\\

\includegraphics[width=0.28\columnwidth,height=0.21\columnwidth]{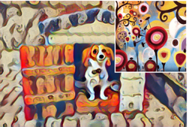}&
\includegraphics[width=0.28\columnwidth,height=0.21\columnwidth]{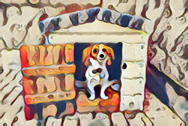}&
\includegraphics[width=0.28\columnwidth,height=0.21\columnwidth]{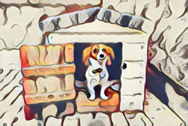}&
\includegraphics[width=0.28\columnwidth,height=0.21\columnwidth]{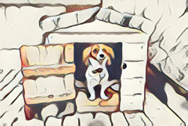}&
\includegraphics[width=0.28\columnwidth,height=0.21\columnwidth]{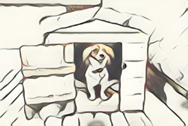}&
\includegraphics[width=0.28\columnwidth,height=0.21\columnwidth]{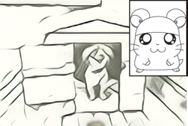}\\

\includegraphics[width=0.28\columnwidth,height=0.21\columnwidth]{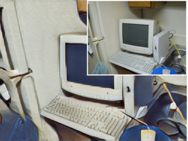}&
\includegraphics[width=0.28\columnwidth,height=0.21\columnwidth]{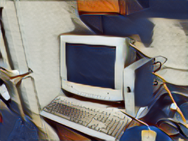}&
\includegraphics[width=0.28\columnwidth,height=0.21\columnwidth]{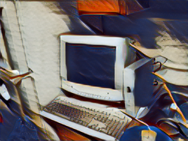}&
\includegraphics[width=0.28\columnwidth,height=0.21\columnwidth]{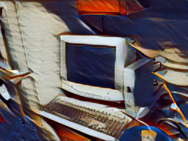}&
\includegraphics[width=0.28\columnwidth,height=0.21\columnwidth]{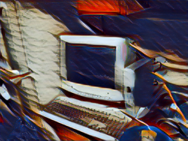}&
\includegraphics[width=0.28\columnwidth,height=0.21\columnwidth]{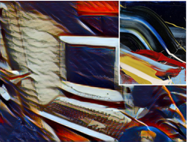}\\
    \end{tabular}
    \small \caption{Interpolations of different styles. The fist two rows show the combination of two styles in the hidden state. In the third row, by feeding the content image to the meta network we can get an identity transformation network, which enables the control of the strength of the other style.}
    \label{figure:interpolations}
\vspace{-0mm}
\end{figure*}

\begin{figure*}[t]
  \centering
  \includegraphics[width=1.7\columnwidth]{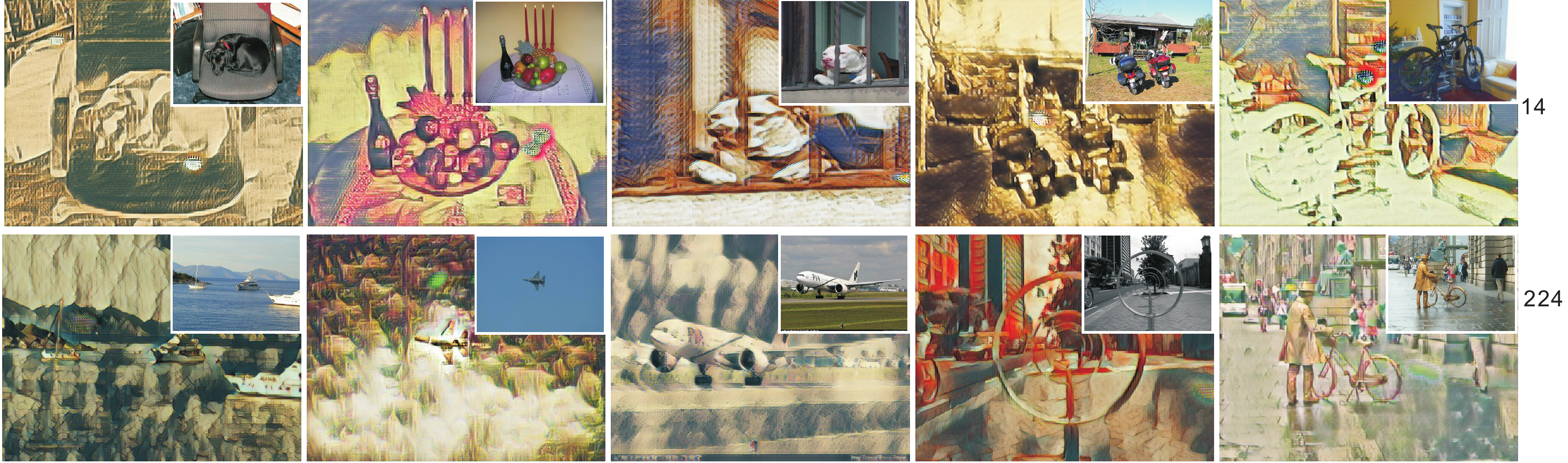}
  \small  \caption{Randomly generated styles at varying hidden state $h \in \{14,224\}$.}
  \label{figure:generate}
\vspace{-0mm}
\end{figure*}


\subsection{Additional experiments}
In this subsection, we further explore the fascinating features of the network manifold from the meta network.

\textbf{Texture Visualization.}
After one forward-propagation, the meta network encodes the style image into the image transformation network. To better understand what kind of target style is learned, we visualize the texture by feeding Gaussian white noise as the content image to the image transformation network to get the uniform texture. Figure~\ref{figure:texture} displays the image-texture pair.

\textbf{Style Interpolation.}
Because of an explicit representation of every network, we could compute the linear interpolation of the hidden states and  get a sensible network, which proves the space continuity in the network manifold. Figure~\ref{figure:interpolations} displays interpolations of hidden states between two real styles. These style images are not part of the training style images. The first and last columns contain images transferred by real style images while the images in between are the results of linear interpolation in hidden states. In the third row, we treat the content image as the style image and get an \texttt{identity transformation network}. By interpolating the hidden state of the identity transformation network and style image transformation network, we can control the strength of the style.

\textbf{Texture Generation.}
Our meta network could also produce sensible texture given random hidden states. Figure~\ref{figure:generate} shows some representative stylish images drawn uniformly from the hidden state. We observe varied light exposure, color and pattern.  The meta network works well, possessing a large diversity of styles.
We compare the effects of the varying dimensions of hidden states. Apparently, the large dimension of hidden states makes better style images. 
\vspace{6mm}
\section{Conclusion}
We introduce the meta network, a novel method to generate the near-optimal network instead of stochastic gradient descent. Especially for neural style transfer, the meta network takes in any given style image and produces a corresponding image transformation network. Our approach provides an efficient solution to real time neural style transfer of any given style.
We also explore the network manifold by operating on  the hidden state in the meta network.
From the experimental results that validate the faster speed of our method with similar performance, we can see that meta networks and direct-mapping for optimization have a successful application to neural style transfer.
\clearpage
{
\small
\bibliographystyle{aaai}
\bibliography{mybib}
}

\end{document}